\documentclass[10pt,twocolumn,letterpaper]{article}

\usepackage{wacv}
\usepackage{times}
\usepackage{epsfig}
\usepackage{graphicx}
\usepackage{amsmath}
\usepackage{amssymb}
\usepackage{hhline}
\usepackage{dsfont}
\usepackage{hyperref}
\usepackage{multirow}
\usepackage{relsize}
\DeclareMathOperator*{\argmin}{arg\,min}
\DeclareMathOperator*{\argmax}{arg\,max}
\mathchardef\mhyphen="2D

\wacvfinalcopy % *** Uncomment this line for the final submission

\def\wacvPaperID{****} % *** Enter the wacv Paper ID here

\ifwacvfinal\fi
\begin{document}

%%%%%%%%% TITLE
\title{Fast Geometrically-Perturbed Adversarial Faces}
% Authors at different institutions
\author{Ali Dabouei, Sobhan Soleymani, Jeremy Dawson, Nasser M. Nasrabadi \\
West Virginia University\\
{\tt\small \{ad0046, ssoleyma\}@mix.wvu.edu, \{jeremy.dawson, nasser.nasrabadi\}@mail.wvu.edu}
% \and
% Second Author \\
}

\maketitle
\ifwacvfinal\thispagestyle{empty}\fi

%%%%%%%%% ABSTRACT
\begin{abstract}
The state-of-the-art performance of deep learning algorithms has led to a considerable increase in the utilization of machine learning in security-sensitive and critical applications. However, it has recently been shown that a small and carefully crafted perturbation in the input space can completely fool a deep model. In this study, we explore the extent to which face recognition systems are vulnerable to geometrically-perturbed adversarial faces. 
We propose a fast landmark manipulation method for generating adversarial faces, which is approximately 200 times faster than the previous geometric attacks and obtains 99.86\% success rate on the state-of-the-art face recognition models. To further force the generated samples to be natural, we introduce a second attack constrained on the semantic structure of the face which has the half speed of the first attack with the success rate of 99.96\%. Both attacks are extremely robust against the state-of-the-art defense methods with the success rate of equal or greater than 53.59\%. Code is available at \href{https://github.com/alldbi/FLM}{\textcolor{blue}{https://github.com/alldbi/FLM}}.

\end{abstract}

%%%%%%%%% BODY TEXT
\section{Introduction}

Machine learning models especially deep neural networks (DNNs) have obtained state-of-the-art performance in different domains ranging from image classification \cite{krizhevsky2012imagenet} to object detection \cite{redmon2017yolo9000} and semantic segmentation \cite{long2015fully}. Despite the excellent performance, it has been shown \cite{szegedy2013intriguing, goodfellow6572explaining} that DNNs are vulnerable to a small perturbation in the input domain which can result in a drastic change of predictions in the output domain. These small perturbations, which are often imperceptible to humans, can transform natural examples into \textit{adversarial examples} that are capable of manipulating high-level predictions of neural networks.

A crucial characteristic of adversarial examples is that they are visually similar to the original samples. This property significantly highlights the vulnerability of DNNs in critical applications where a carefully crafted adversarial example may remain benign to the human eye while targeting several machine learning models. For instance, autonomous vehicles may be misled by traffic signs constructed by an adversary to deceive machine learning methods, while the same sign may seem natural to human drivers \cite{kurakin2016adversarial}.
\begin{figure}
\begin{center}
\includegraphics[scale=.45]{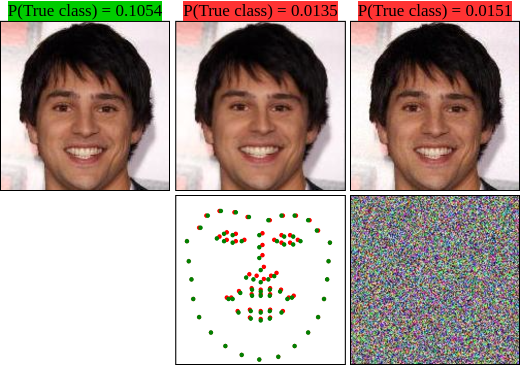}
\end{center}
   \caption{Comparison of the proposed attack to an intensity-based attack. First column: the ground truth image, which is correctly classified. Second column: the spatially transformed adversarial image wrongly classified and the corresponding adversarial landmark locations computed by our method. Third column: the adversarial image wrongly classified and the corresponding perturbation generated by the fast gradient sign method \cite{goodfellow6572explaining}. The proposed method leads to natural adversarial faces which are clean from additive noise. } 
\label{fig:compare}
\end{figure}

Most of the attack methods developed in the previous works \cite{gu2014towards, goodfellow6572explaining, moosavi2016deepfool} are intensity-based attacks, as they directly manipulate the intensity of input images to fool the target model. Intensity-based attacks are computationally cheap and can prosper from a low-cost similarity constraint by adopting an $\ell_p\mhyphen norm$ to force the generated examples to be similar to the benign samples. Since perturbations for neighborhood pixels are computed independently, adversarial examples generated using intensity-based attacks often have high-frequency components that can be used as a measure to detect and remove them \cite{liao2018defense}.
On the other hand, the $\ell_p\mhyphen norm$ is not a perfect measure for perceptual similarity since it is sensitive to spatial transformations \cite{johnson2016perceptual}. 
 For instance, a small rotation, translation, or scale variation in the input image, results in a drastic change of similarity.   
These limitations restrict intensity-based attacks from incorporating spatial perturbations. Recently, Xiao \etal \cite{xiao2018spatially} proposed a novel method of generating adversarial examples by spatially transforming natural images. Spatial transformations provide a convenient way of incorporating neighborhood information through interpolation.

From the defensive perspective, face recognition systems can be divided into two different types, active and passive. In the active type, the model processes online face images from devices such as surveillance or access control cameras to identify the captured face. Therefore, the model has a limited amount of time to examine whether the input image is natural or not. In the passive face recognition, individuals submit a digital or hard copy photo to register their identity in a system for future identification. The attacker can submit an adversarial face image that prevents the system from recognizing the malicious ID in the future. In such a case, the defense algorithm has unlimited time to examine the gallery images. Hence, attacking passive face recognition systems is more challenging than attacking active systems. However, if the attack on the passive face recognition system is successful, the attacker may obtain a long-term immunity against the identification system.      

This study explores the extent to which passive face recognition systems are vulnerable to spatially transformed adversarial examples. Inspired by \cite{xiao2018spatially}, we propose a novel and fast method of generating adversarial faces by altering the landmark locations of the input images. The resulting adversarial faces completely lie on the manifold of natural images, which makes it extremely hard for defense methods to detect them even by a novelty detector \cite{wang2017safer}.  The contributions of this paper are as follows:

\begin{itemize}
    \item We have demonstrated that the prediction of a face recognition model has a linear trend around the actual value of the landmark locations of the input face image.
    
    \item We have introduced a fast method of generating adversarial face images, which is approximately 200 times faster than the previous geometry-based attacks which use L-BFGS optimization.
    
    \item We have developed a structure-constrained attack that manipulates face landmarks based on the semantic regions of the face. 
    
    \item We have demonstrated that constraining the attack to preserve the natural structure of faces greatly increases the robustness of the method against the state-of-the-art defense algorithms.   
\end{itemize}

\section{Related Work}

Recent advances in technology have led to the generation of large datasets and powerful computational resources that made it possible to train deeper learning models. These models outperformed traditional methods in different areas ranging from signal processing to action recognition. Despite the spectacular performance, Szegedy \etal \cite{szegedy2013intriguing} showed that a small perturbation in the input domain can fool a trained classifier into making a wrong prediction confidently. In this section, we first review the literature on intensity-based and geometry-based attacks. Then we explore the background of adversarial examples for the face recognition systems. 
\subsection{Intensity-Based Attacks}

\begin{figure*}[t]
\begin{center}
\includegraphics[scale=.10]{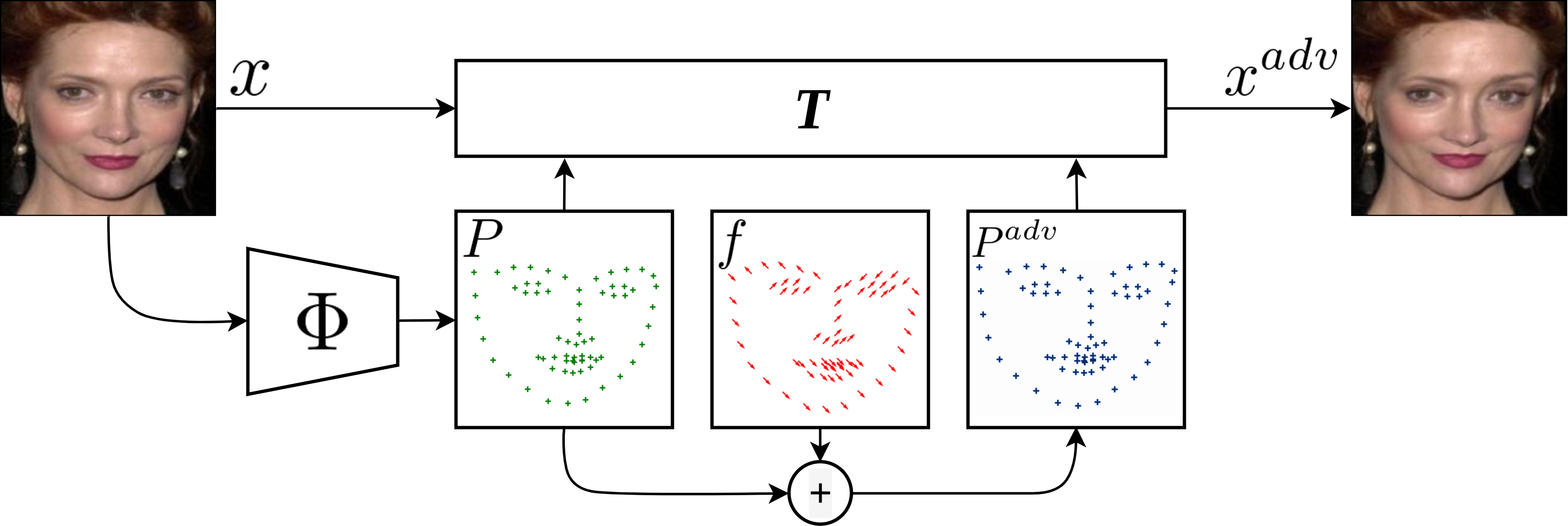}
\end{center}
   \caption{
   The proposed method optimizes a displacement field $f$ to produce adversarial landmark locations $P^{adv}$. The spatial transformation $T$ transforms the input sample to the corresponding adversarial image $x^{adv}$ such that $\Phi(x^{adv}) = \Phi(x) + f$, and a state-of-the-art face recognition model $g$ miss-classifies the transformed image $x^{adv}$.  }
\label{fig:main}
\end{figure*}

Algorithms for generating adversarial examples can be categorized by the perturbation type. Most of the previously proposed methods are intensity-based attacks, as they directly try to manipulate the intensity of the input sample. Szegedy \etal \cite{szegedy2013intriguing} used a box-constrained L-BFGS \cite{liu1989limited} to generate some of the very first adversarial examples. Despite the high computational cost, their method was able to fool many networks trained on different inputs.

Goodfellow \etal \cite{goodfellow6572explaining} proposed a fast and efficient intensity-based attack called the Fast Gradient Sign Method (FGSM) and showed that the prediction of a deep leaning model has a linear trend around the saddle point of the input sample. Hence, they used the sign of the gradient of the classification loss with respect to the input sample as the perturbation to manipulate the intensity of the benign examples. This provides a fast and effective single-step attack. Although they select a small coefficient for the amplitude of the gradient sign
to make the perturbation imperceptible, such a noisy pattern can facilitate the process of defending against it \cite{liao2018defense, liang2017detecting}. Various extensions to intensity-based attacks have been developed to explore the vulnerability of machine learning models. To increase the effectiveness of the attack, Rozsa \etal \cite{rozsa2016adversarial} proposed to use the actual gradient value instead of the gradient sign used in FGSM \cite{goodfellow6572explaining}. Also, several iterative methods are developed to improve the robustness of single-step attacks against defenses, including the iterative version of FGSM \cite{dong2018boosting, kurakin2016adversarial}. 

Papernot \etal \cite{papernot2016limitations} proposed the use of the Jacobian matrix of the prediction of classes concerning the input sample to generate \textit{Jacobian-based Saliency Map Attack} (JSMA). JSMA reduces the number of pixels that are needed to be changed during the attack by calculating a saliency map of the most important pixels in the input space. Carlini and Wagner \cite{carlini2017towards} modified the JSMA by changing the target layer used in the algorithm to compute the Jacobian matrix. They reported the adversarial success rate of $97\%$ by modifying less than $5\%$ of pixels in the input samples. However, saliency-based methods are computationally expensive due to the greedy search for finding the most significant areas in the input sample. 

Almost all intensity-based attacks add high-frequency components to the input samples and use an $\ell_p\mhyphen norm$ constraint to control the amount of distortion. However, the $\ell_p\mhyphen norm$ is not a perfect similarity measure and does not guarantee that the adversarial samples lie on the same manifold as the natural samples. This increases the vulnerability of intensity-based attacks, especially in the passive applications where the agent has unlimited time to assess the legitimacy of the inputs.

\subsection{Geometry-Based Attacks}

Recently, Xiao \etal \cite{xiao2018spatially} proposed stAdv attack in which they generate adversarial examples by spatially transforming benign images. For this purpose, they define a flow field $f$ for all pixel locations in the input image. The corresponding location of a pixel in the adversarial image can be computed by the displacement field. Since the displacement field can hold fractional values, they use a differentiable bilinear interpolation \cite{jaderberg2015spatial} to overcome the discontinuity problem. Furthermore, they added the sum of the total displacement of any two adjacent pixels to the main loss function to control the amount of distortion introduced by the displacement field. However, optimizing a flow field for all pixels in an image produces a highly non-convex cost function. They used the L-BFGS \cite{liu1989limited} with a linear backtrack search to find the optimal flow field $f^*$. 
Such a computationally expensive optimization is the critical limitation of this method. 

\subsection{Attacking Face Recognition}
All of the previously proposed attack methods can be adopted for face recognition models, but the approach is highly dependent on the type of the face recognition model. For active face recognition, it has been shown that putting on enormous amounts of makeup \cite{harvey2017cv} or wearing carefully crafted accessories \cite{sharif2016accessorize} can conceal the identity of the attacker. However, wearing heavy makeup or overt accessories may draw attention and increase the chance of defense against the attack.

For passive face recognition, Goel \etal \cite{goel2018smartbox} examined several intensity-based attacks and showed that they are extremely successful in fooling face recognition systems. However, the noisy structure of the perturbation makes these attacks vulnerable against conventional defense methods such as quantizing \cite{liang2017detecting}, smoothing \cite{goel2018smartbox} or training on adversarial examples \cite{szegedy2013intriguing}.  

\subsection{Defense Methods}
Since the introduction of adversarial examples, many approaches have been proposed to detect and mitigate these threats. Current defenses against adversarial attacks consist of two main approaches which modify either the model \cite{goodfellow6572explaining, tramer2017ensemble, madry2017towards} or the input before feeding to the model \cite{das2017keeping, wang2016learning}. The most successful group of defenses to date are methods based on modifying the model, especially by using adversarial training \cite{madry2017towards}.The adversarial training uses the adversarial examples during the training phase to make the model robust against the attack. Goodfellow \etal \cite{goodfellow6572explaining} proposed to utilize FGSM to generate adversarial examples and use them to train the model to provide robustness against adversarial examples. Later in Section \ref{sec:underattack}, we use this method followed by ensemble adversarial training \cite{tramer2017ensemble} and projected gradient descent \cite{madry2017towards} to examine the performance of our attacks under these state-of-the-art defenses.

\section{Approach}  
Here we first briefly describe the problem of generating adversarial examples. We then define a face transformation model based on the landmark locations in Section \ref{sec_landmarkbased}. We continue by presenting a landmark-based attack in Section \ref{sec:fastlandmarkmanipulation} and developing a structural constraint in Section \ref{sec:semanticgrouping}.

\subsection{Problem Definition}
For the process of generating adversarial faces, we assume that the victim face recognition model is a well-trained classifier $g: x \rightarrow y$ over $N_c$ different classes, that predicts a vector of classification scores $y \in \mathbb{R}^{N_c}$, given an input face image $x \in [0, 1]^{H\times W \times 3}$ with spatial size $H\times W$. We consider the white-box scenario where the attacker has full knowledge about the model and its prediction. The attacker tries to manipulate a benign face image $x$ from class $c$ in a way that the face recognition model miss-classifies the resulting adversarial face image $x^{adv}$. 

\subsection{Landmark-Based Face Transformation}
\label{sec_landmarkbased}
Let  $\Phi$ be a landmark detector function that maps the face image $x$ to a set of $k$ 2D landmark locations $P=\{p_1,\ldots, p_k\}$, $p_i=(u_i, v_i)$. We assume $p_i^{adv}=(u_i^{adv}, v_i^{adv})$ is the transformed version of $p_i$, and defines the location of the $i$-th landmark in the corresponding adversarial face image $x^{adv}$. To manipulate the face image based on $P$, we define the per-landmark flow (displacement) field $f$ to produce the location of the corresponding adversarial landmarks. For the $i$-th landmark $p_i^{adv}=(u_i^{adv}, v_i^{adv})$, we optimize the spatial displacement vector $f_i = (\Delta u_i, \Delta v_i) $\footnote{We assume that 2D coordinates are independent. So in the rest of the paper, all operations on coordinates are element-wise.}. The adversarial landmark $p_i^{adv}$ can be obtained from the original landmark $p_i$ and the displacement vector $f_i$ as:

\begin{equation}
\begin{split}
    &~~~~~~~~~~~~~p_i^{adv} = p_i + f_i,\\ &(u_i^{adv}, v_i^{adv}) = (u_i+\Delta u_i, v_i+\Delta v_i ).
\end{split}
    \label{eq_field}
\end{equation}

Contrary to \cite{xiao2018spatially}, which estimates the displacement field $f$ for all pixel locations in the input image, the displacement field $f$ in the proposed method is only defined for $k$ landmarks. In a real-world application, especially face recognition problems, $k$ is notably small compared to the number of pixels in the input image. As a result, it is possible to use conventional spatial transformations to transform the input image. Consequently, limiting the number of control points reduces the distortion introduced by the spatial transformation. The resulting adversarial face image is the transformed version of the benign face image using the transformation T as follows:

\begin{equation}
x^{adv} = T(P, P^{adv}, x),
    \label{eq_ut}
\end{equation}
where $T$ is the spatial transformation that maps the source control points $P$ to the target control points $P^{adv}$. Note that $x^{adv}$ is differentiable with respect to the landmark locations and the input image. 

\begin{figure}[t]
\begin{center}
\includegraphics[scale=.25]{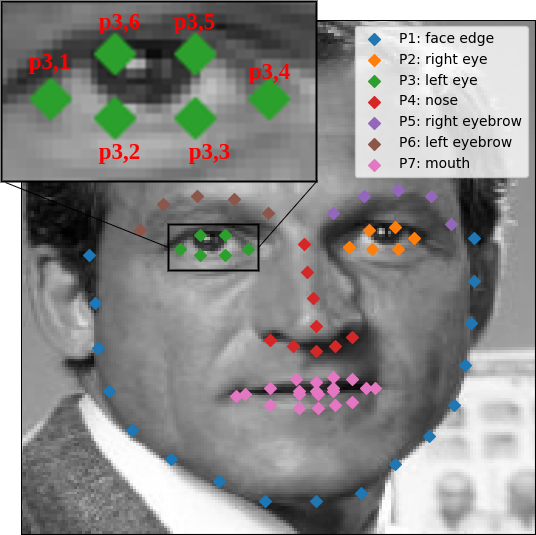}
\end{center}
   \caption{Grouping face landmarks based on semantic regions of the face.}
\label{fig:lnd_map}
\end{figure}

% \subsection{Fast Geometric Perturbation}
\subsection{Fast Landmark Manipulation}
\label{sec:fastlandmarkmanipulation}
It has been shown \cite{irfanoglu20043d} that landmark locations in the face image provide highly discriminative information for face recognition tasks. Indeed, we experimentally show this in Section \ref{sec:interpolate} that even learning based face recognition systems discriminate face identities based on extracting the relative geometric features. More specifically, the predictions of face recognition systems are highly linear around the original landmark locations of the face image. This property allows the direct employment of the gradient of the prediction in a face recognition model to geometrically manipulate benign faces. 

We use the gradients of the prediction with respect to the location of landmarks to update the displacement field $f$. For this purpose, we first define a standard for the correct prediction, and then we use it to compute the formulation of the attack. As a measure of correct classification, we select the same \textit{softmax cost} used in \cite{sharif2016accessorize, parkhi2015deep}. 
Given an input $x$, a one-hot label vector $y_c$ corresponding to class $c$ and a vector of classification score $g(x)$ from the victim classification model, we define the \textit{softmaxcost} as:

\begin{equation}
\label{eq_softmax}
\mathlarger{J}\big(g(x), c\big) = -\log\bigg(\frac{e^{y_c^Tg(x)}}{\sum_{n=1}^{N_c}e^{y_n^Tg(x)}}\bigg),
\end{equation}
where $N_c$ is the number of classes. Besides, we define a boundary for the amount of displacement to prevent the model from generating distorted face images. 
Inspired by \cite{xiao2018spatially}, we develop $L_{flow}$ to constrain the displacement field $f$ as follows:

\begin{equation}
\label{eq_lflow}
L_{flow}(f) = \frac{1}{k} \sum_{i=1}^{k}({\Delta u_i}^2 + {\Delta v_i}^2).
\end{equation}

Having the measure for the correct classification, and the term for bounding the displacement field, we define the total loss for generating adversarial faces as:

\begin{equation}
\label{eq_total_loss}
\begin{split}
L_t(P, P^{adv}, x, c_x)=\mathlarger{J}\Big(\mathlarger{g}\big(T(P, P^{adv}, x)\big), c_x\Big)\\ - \lambda_{flow} \mathlarger{L}_{flow}(P^{adv}-P),
\end{split}
\end{equation}
where $\lambda_{flow}$ is a positive coefficient used to control the magnitude of the displacement.
The attacker can generate geometric adversarial perturbations by finding the $f^*$ as:

\begin{equation}
\label{eq_optdodging}
f_{}^*=\argmax_{f} L_t(P, P^{adv}, x, c_x).
\end{equation}

As we show in Section \ref{sec:interpolate}, the prediction of face recognition models is highly linear around the ground truth location of the face landmarks; therefore, we use the direction of the gradient of the prediction (same as FGSM \cite{goodfellow6572explaining}) to find the landmark displacement field $f$ in an iterative manner. The $t$-th optimization step for finding $f$ using FGSM is:
\begin{equation}
\label{eq_fgsm5}
\begin{split}
f^{(t+1)} = f^{(t)} + ~~~~~~~~~~~~~~~~~~~~~~~~~~~~~~~~~~~~~~~~~~~~~~~~~~~~~~~~~~\\ \epsilon \, {sign}(\nabla_{P^{adv(t)}}L_{t}(P, P^{adv(t)}, x, c_x)),~~~~~~
\end{split}
\end{equation}
where $P^{adv(t)} = P + f^{(t)}$. We refer to this as the \textit{fast landmark manipulation method} (FLM) for generating adversarial faces. Figure \ref{fig:main} shows an overview of the method.

\subsection{Semantic Grouping of Landmarks}
%So far we 
\label{sec:semanticgrouping}
In the previous section, we developed a model to generate face images based on manipulating the landmark information. Although this method is fast and computationally cheap, it has a limitation that should be addressed. In Equation \ref{eq_fgsm5}, we use the gradients of the classification loss with respect to the landmark locations to update the displacement field for generating the adversarial face images. These gradients can have any direction in the 2D coordinate space. As a result, multiple updates of the displacement field $f$ can severely distort the generated adversarial images. To prevent this issue, we adopt the total $\ell_2{-}norm$ of the displacement field $f$ as an additional loss. However, our model computes the displacement field $f$ for a significantly small number of locations in the input image, so limiting the size of $f$ can reduce the effectiveness of the attack. 

To overcome this limitation, we propose to semantically group landmarks and manipulate the group properties instead of perturbing each landmark. Consequently, the total structure of the face will be preserved. This consideration allows us to increase the total amount of displacement and, as a result, extremely increases the effectiveness of the attack. We break down the set of landmarks $P$ into $m$ semantic groups $P_i, i \in \{1,\ldots, m\}$, and $p_{i, j}$ denotes the $j$-th landmark in the $i$-th group which has $n_i$ landmarks. These groups are formed based on their semantic regions in the face, such as left eye, right eye, mouth, \etc. Figure \ref{fig:lnd_map} shows a sample grouping of face landmarks used in this study. We define a flow field vector for each of the groups by means of a translation and a scale variable that will apply to all elements in the group. For the face regions, a rotation is not of interest because it is not natural to have a face with a rotated mouth or nose.

Let $P_i$ be the $i$-th landmark group \eg all landmarks of the nose. To scale these landmarks, we define the scaling tuple $\alpha_i=(\alpha_{u_i}, \alpha_{v_i})$ where $\alpha_{u_i}$ and $\alpha_{v_i}$ are the horizontal and vertical scaling parameters respectively. To translate the landmarks, we define the translation tuple $\beta_i=(\beta_{u_i}, \beta_{v_i})$ where $\beta_{u_i}$ and $\beta_{v_i}$ are the translation parameters for the horizontal and vertical axes respectively. The location of the corresponding landmarks in the adversarial image can be computed as:

\begin{equation}
P_i^{adv} = \alpha_i (P_i-\overline{p_i}) + \beta_i,
    \label{eq_scaleandtranslate}
\end{equation}
where $\overline{p_i}{=}\frac{1}{n_i}\sum_{j=1}^{n_i}p_{i,j}$ is the average location of all landmarks in the group $P_i$. We subtract the average of the group from each landmark location in the group before scaling to force each part of the face to be scaled regarding its center. 

We choose $\alpha_i$ and $\beta_i$ such that they minimize the square error of $P_i^{adv}$ between Equation \ref{eq_field} and Equation \ref{eq_scaleandtranslate} as:

\begin{equation}
    \argmin_{\alpha_i, \beta_i} \dfrac{1}{n_i} \sum_{j=1}^{n_i} \big(\alpha_i({p}_{i,j}-\overline{p_i })+\beta_i-{p}_{i,j} - f_{i,j} \big)^2.
    \label{eq_2a}
\end{equation}
Solving Equation \ref{eq_2a} results in the closed-form solutions for the $\alpha_i$ and $\beta_i$ as:

\begin{equation}
    \alpha_i = \frac{\sum_{j=1}^{n_i} (p_{i,j}-\overline{p_i})(p_{i,j}+f_{i,j})}{\sum_{j=1}^{n_i}(p_{i,j}-\overline{p_i})^2},
    \label{eq_3a}
\end{equation}

\begin{equation}
    \beta_i = \overline{p_i\vphantom{f_i}} + \frac{1}{n_i} \sum_{j=1}^{n_i}f_{i,j}.
    \label{eq_3b}
\end{equation}
We modeled the effect of the displacement field $f_{i,j}$ for each group of landmarks as a scaling and a translation function. While Equation \ref{eq_fgsm5} optimizes $f$, we use Equations \ref{eq_3a} and \ref{eq_3b} to calculate the corresponding set of scale tuples $\{\alpha_1,\ldots, \alpha_7\}$ and translation tuples $\{\beta_1,\ldots, \beta_7\}$. We refer to this as the \textit{grouped fast landmark manipulation method} (GFLM) for generating adversarial faces.

\section{Experiments}

\begin{figure*}[t]
\begin{center}
\includegraphics[scale=.3]{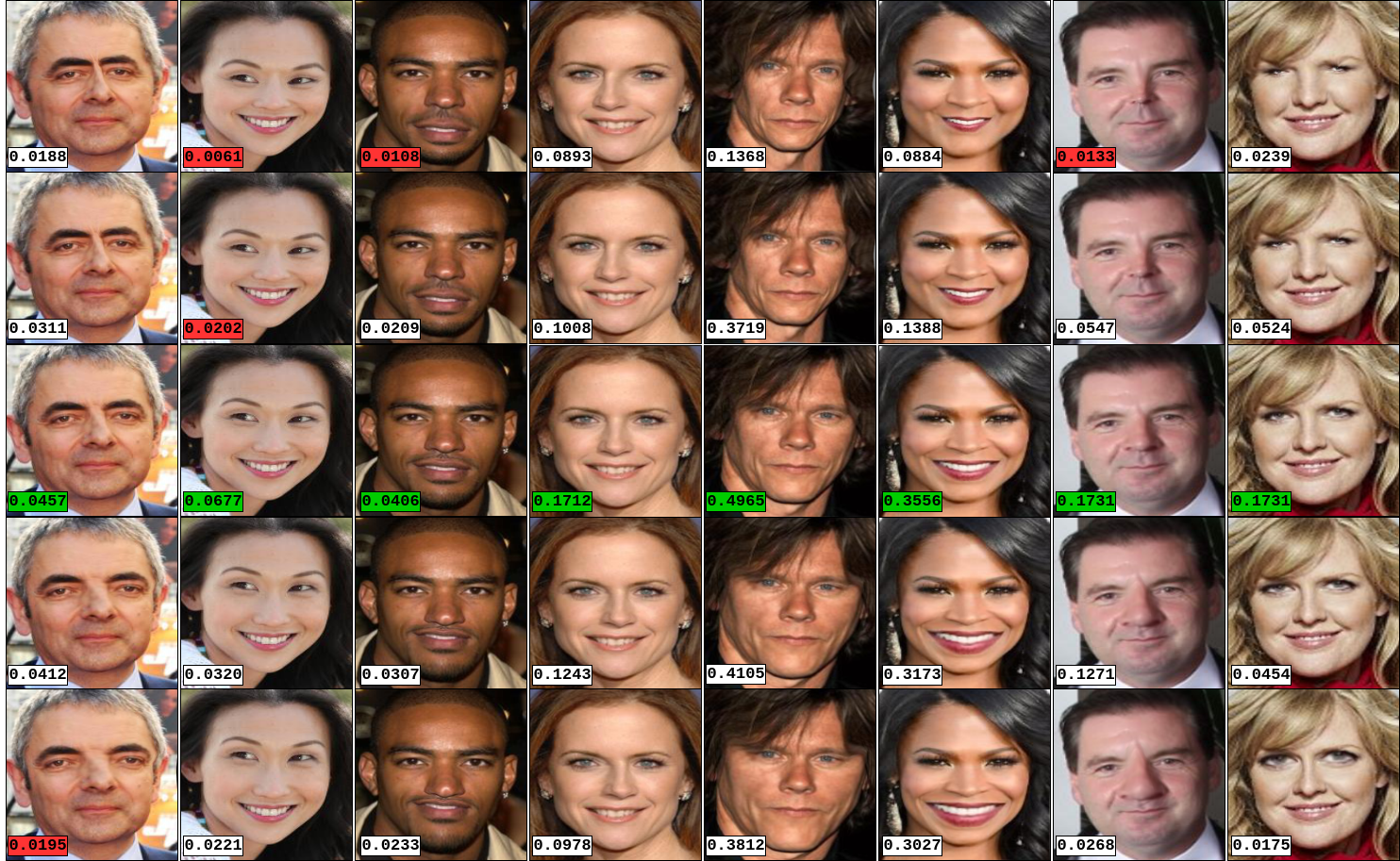}
\end{center}
   \caption{Examples of linearly interpolating face properties. Each column from Left to right shows examples of interpolating one of the eight geometric variables of the face structure described in Section \ref{sec:interpolate}. The probability of the true class is depicted on the bottom left corner of samples. The green color specifies the face image with the maximum probability of belonging to the true class. The red color shows the incorrectly classified face images. }

\label{fig:face_samples}
\end{figure*}

We first describe the implementation details in Section \ref{exp_setup}. Then we investigate how landmark information influences the prediction of a face classifier in Section \ref{sec:interpolate}.
We evaluate the performance of the proposed attacks in the white-box scenario in Section \ref{sec:whitebox} and conclude the experiments by measuring and comparing the performance of our attacks under several defense methods in Section \ref{sec:underattack}.

\subsection{Implementation Details}
\label{exp_setup}

To evaluate the performance of the proposed method in the white-box scenario, we use the face recognition model developed by Schroff \etal \cite{schroff2015facenet} that obtained the state-of-the-art results on the Labeled Faces in the Wild (LFW) \cite{huang2008labeled} challenge as the victim model. We train two instances of the model\footnote{\url{https://github.com/davidsandberg/facenet}} on two datasets of face images. The first instance is trained to recognize 9,101 celebrities from the VGGFace2 dataset \cite{cao2018vggface2} with more than 3.3M training images and the average of 360 images per subject.
The second instance is trained on the CASIA-WebFace \cite{yi2014learning} dataset which consists of more than 494,000 face images and 10,575 unique IDs. For extracting the landmark information of the input face images, we use the Dlib \cite{king2009dlib} landmark detector which predicts the 2D coordinates for 68 landmarks. We divide landmarks based on five facial regions as: 1) $P_1{:}$ \textit{jaw}, 2) $P_2{:}$ \textit{right eye and eyebrow}, 3) $P_3{:}$ \textit{left eye and eyebrow}, 4) $P_4{:}$ \textit{nose}, and 5) $P_5{:}$  \textit{mouth}. The number of landmarks in each group is as: $\{ n_1{=}17, n_2{=}11, n_3{=}11, n_4{=}9, n_5{=}20\}$. Figure \ref{fig:lnd_map} demonstrates a similar grouping of landmarks.

We opt to use the thin plate spline \cite{bookstein1989principal} (TPS) to cover a broad range of spatial transformations that are capable of locally manipulating face images. TPS has $2(k+3)$ parameters for mapping $k$ source landmarks $P$ to their corresponding $P^{adv}$. We first scale coordinates to lie inside the range $[-1, 1]^2$ where $(-1, -1)$ is the top left corner and $(1, 1)$ is the bottom right corner of the image. We assume all coordinates are continuous values since TPS has no restriction on the continuity of the coordinates because of the differentiable bilinear interpolation \cite{jaderberg2015spatial}.    

We set the value of $\lambda_{flow}$ for the FLM attack to 100. For the GFLM we do not set any limit for the amount of displacement since the structural condition developed in Section \ref{sec:semanticgrouping} is enough to preserve the similarity of the generated adversarial examples. Therefore, we set $\lambda_{flow}$ for the GFLM attack to zero.   
To further condition the model to generate realistic faces, we perform an extra modification for the symmetric parts, such as eyes. We set an equal scale and an equal vertical position for these parts. Other conditions can be applied by slightly changing Equation \ref{eq_2a}. For example, instead of manipulating the horizontal location of the eyes independently, one can change the horizontal distance between them to preserve the natural symmetry.  

\begin{figure*}[t]
\begin{center}
\includegraphics[scale=.25]{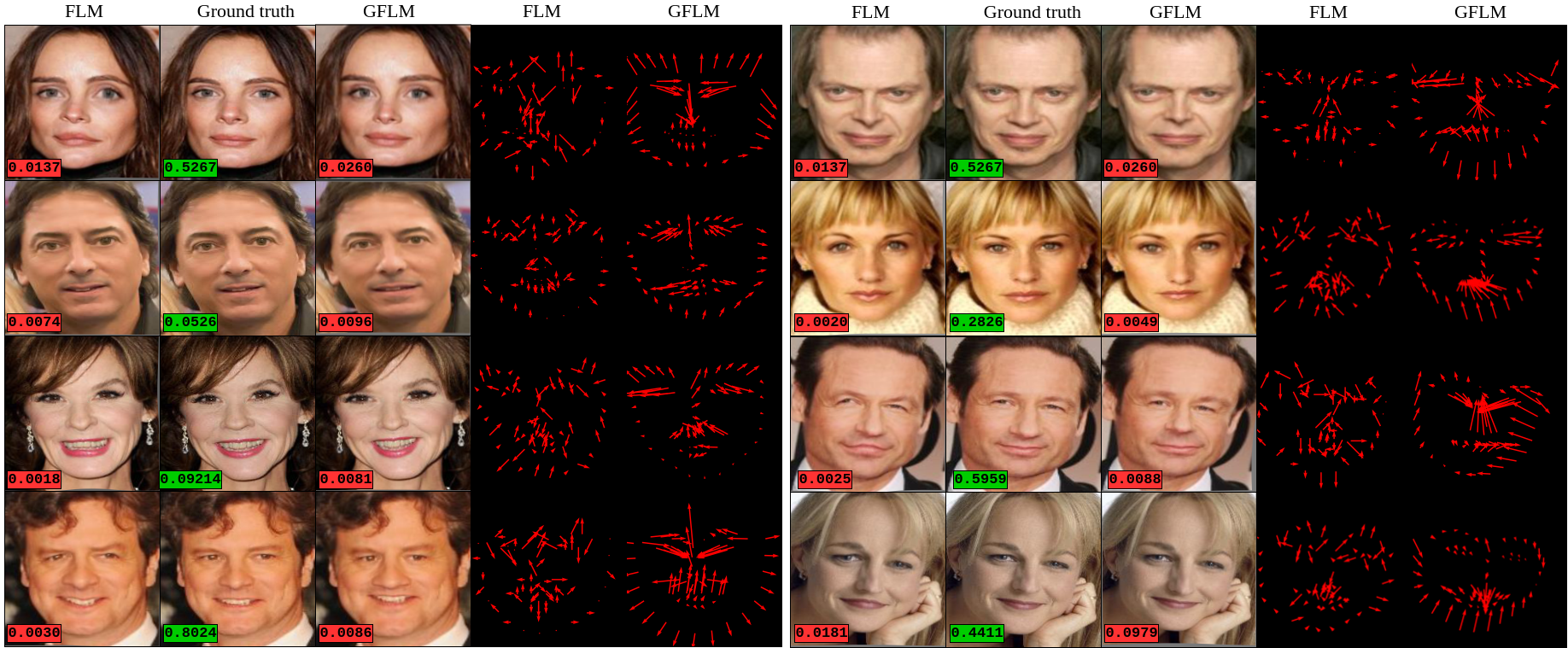}
\end{center}
   \caption{Examples of the adversarial faces generated using FLM and GFLM. For each subject, five images are shown including the original face image (middle face), the result of GFLM (right face), the result of FLM (right image), displacement field $f$ for GFLM (left field) and displacement field $f$ for FLM (right field). Tags on the bottom left of images show the probability of the true class. Green and red tags denote the correct and incorrect classified samples respectively. }
\label{fig:results}
\end{figure*}

\subsection{Interpolated Perturbation}

\begin{figure}[]
\begin{center}
\includegraphics[scale=0.42]{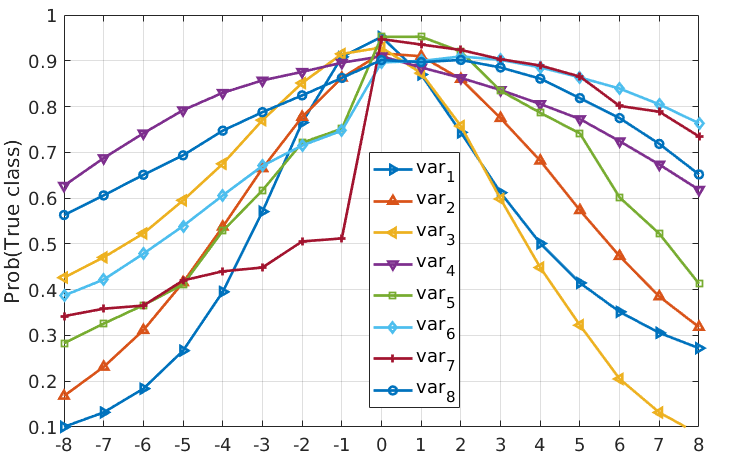}
\end{center}
   \caption{Normalized probability of the true classes based on interpolating the eight variables of face geometry defined in Section \ref{sec:interpolate}.}
\label{fig:plots}
\end{figure}

\label{sec:interpolate}

The geometry of the face is unique and provides highly discriminative information for face recognition. In this section, we perform an experiment to evaluate how spatially manipulating the face regions affects the performance of a face recognition system.
We extract landmarks for all faces in the CASIA-WebFace \cite{yi2014learning} dataset and define eight variables based on the geometric properties of the face regions. The first four variables are the translation-based variables which are: 1) horizontal distance between the eyes and eyebrows, 2) vertical location of the eyes and eyebrows, 3) horizontal location of the nose,  4) horizontal location of the mouth. The second set of variables are the scale-related variables and are as follows: 5) scale of the jaw, 6) scale of the mouth, 7) scale of the nose, and 8) scale of the eyes.
We interpolated each of these variables independently to measure the influence of each on the performance of the face recognition model. Figure \ref{fig:face_samples} shows several examples of the interpolation. 

We calculate the prediction of the true class for faces which are correctly classified and their manipulated versions. The predictions are averaged over all the ID's to investigate how manipulating face parts affects the predicted probability of the class. Figure \ref{fig:plots} shows the final averaged values for the predictions. As it is shown, the global maximum of the model's prediction for a sample face is around the ground truth value of the positions and the scales. These results confirms that the geometry of the face contains highly discriminative information for face recognition. Indeed, the prediction of a face recognition model has a linear characteristic around the actual size and location of face regions and enables us to directly use the gradient of the prediction to manipulate landmark locations.

\begin{table*}[h]
\begin{center}

\begin{tabular}{c|c|cccc|cccc|ccc}
\hline
\multirow{2}{*}{\textit{\#}} & \multirow{2}{*}{\begin{tabular}[c]{@{}c@{}}Face\\ Region\end{tabular}} & \multicolumn{4}{c|}{\textbf{FLM}} & \multicolumn{4}{c|}{\textbf{GFLM}} & \multicolumn{3}{c}{\textbf{stAdv \cite{xiao2018spatially}}} \\ \cline{3-13} 
 &  & $\overline{n}$ & SR(\%) & pT & T(s) & $\overline{n}$ & SR(\%) & pT & T(s) & SR(\%) & pT & T(s) \\ \hline
1 & Eyebrows & 9.6 & 61.92 & 0.0156 & 0.473 & 21.1 & 79.45 & 0.0159 & 1.019 &  &  &  \\ \cline{1-10}
2 & Eyes & 11.5 & 55.59 & 0.0169 & 0.511 & 18.6 & 83.06 & 0.0156 & 0.781 &  &  &  \\ \cline{1-10}
3 & Nose & 8.8 & 82.50 & 0.0139 & 0.495 & 10.7 & 89.37 & 0.0149 & 0.575 &  &  &  \\ \cline{1-10}
4 & Mouth & 10.4 & 49.27 & 0.0152 & 0.455 & 20.2 & 77.13 & 0.0165 & 0.868 &  &  &  \\ \cline{1-10}
5 & Jaw & 8.7 & 12.54 & 0.0180 & 0.420 & 37.6 & 49.26 & 0.0161 &  1.804 &  &  &  \\ \hline
6 & All & \textbf{2.8} & 99.86 & \textbf{0.0072} & \textbf{0.126} & 5.2 & \textbf{99.96} & 0.0120 & 0.254 & 99.18 & 0.0166 & 27.177 \\ \hline
\end{tabular}
\end{center}
\caption{Comparing results of the proposed attacks to stAdv \cite{xiao2018spatially} and exploring the influence of different regions of the face on our attacks. In each experiment, the average number of iterations ($\overline{n}$), the success rate of the attack(SR), the average final probability of the true class (pT), and the average time of the attack are shown.}
\label{tab:white-box}
\end{table*}

\begin{table}[h]
\begin{center}
\begin{tabular}{ccccc}
  \hline
\multicolumn{1}{c}{\textit{Defense}} & FGSM \cite{goodfellow6572explaining} & stAdv \cite{xiao2018spatially} & FLM & GFLM \\ \hhline{=====}
\multicolumn{1}{c}{Adv. \cite{goodfellow6572explaining}} & 19.12 & 36.96 & 54.79 & \textbf{62.03} \\ \hline
\multicolumn{1}{c}{Ens. \cite{tramer2017ensemble}} & 16.27 & 33.80 & 53.59 & \textbf{61.84} \\ \hline
\multicolumn{1}{c}{PGD \cite{madry2017towards}} & 18.95 & 39.15 & 55.65 & \textbf{67.43} \\ \hline
\end{tabular}
\end{center}
\caption{Comparing the success rate of the proposed FLM and GFLM attacks to FGSM \cite{goodfellow6572explaining} and stAdv \cite{xiao2018spatially} attacks under the state-of-the-art adversarial training defenses. }
\label{tab:resultsunderattack}
\end{table}

\subsection{White-Box Attack}
\label{sec:whitebox}
We evaluate the performance of both proposed methods of FLM and GFLM for the white-box attack scenario on the CASIA-WebFace \cite{yi2014learning} dataset. We define six experiments to investigate the importance of each region of the face in the FLM and GFLM attack methods. In the first five experiments, we evaluate the performance of the attacks on each of the five main regions of the face including 1) eyebrows, 2) eyes, 3) nose, 4) mouth and 5) jaw. In the last experiment, we evaluated the performance of attacks using all five regions of the face. Also, in Experiment 6, we compare the performance and speed of the proposed methods to the method developed by \cite{xiao2018spatially} in which the displacement field $f$ is defined for all pixels in the input image. All the experiments are conducted on a PC
with 3.3 GHz CPU and NVIDIA TITAN X GPU. Table \ref{tab:white-box} shows the results for all the six experiments.

From the results, we observe that both the FLM and GFLM are generating powerful adversarial face images that fool the classifier for more than 99.86\%  of the samples. An important point is the computation time of these algorithms. The average time of generating adversarial faces for the FLM and GFLM is 125 and 254 milliseconds respectively, which is significantly shorter than the computation time of stAdv \cite{xiao2018spatially}, which is 27.177 seconds on average. Indeed, the FLM is 215 and GFLM is 106 times faster than stAdv \cite{xiao2018spatially} method. Furthermore, we described in Section \ref{sec:semanticgrouping} that the FLM can generate faces with spatial distortions, and grouping the landmarks in the GFLM overcomes this problem. Figure \ref{fig:results} demonstrates several examples of the adversarial faces generated by the FLM and GFLM.

\subsection{Performance Under Attacks}
\label{sec:underattack}
To evaluate the performance of the proposed methods under attack, we repeat the sixth experiment in the previous section. For this purpose, we use three state-of-the-art defenses of FGSM adversarial training \cite{goodfellow6572explaining}, PGD adversarial training \cite{madry2017towards}, and ensemble adversarial training \cite{tramer2017ensemble}. We compare the performance of our attacks to FGSM \cite{goodfellow6572explaining} and stAdv \cite{xiao2018spatially}. Results are shown in Table \ref{tab:resultsunderattack}.   
The FLM and GFLM attacks are extremely robust against adversarial training compared to FGSM \cite{goodfellow6572explaining} and stAdv \cite{xiao2018spatially} because they are targeting the most important locations in the benign samples using geometric perturbations. These locations contain the most critical discriminative information that a face recognition model needs to identify an individual. Defenses based on adversarial training use the intensity-based attacks to generate samples for training the model. However, the generated samples do not lie on the manifold of natural images due to the slight change of intensity of all pixels in the input image. Furthermore, the GFLM is more robust against defenses than the FLM since samples generated by the GFLM are conditioned to have the similar structure as a natural face.

\section{Conclusion}
In this paper, we introduced a novel method for generating adversarial face images by manipulating landmark locations of the natural images. Landmark locations contain the most discriminative information that is needed to identify an individual. Therefore, manipulating landmark locations is a strong way to change the prediction of a face recognition system. We experimentally showed that the prediction of a face recognition model has a linear trend around the parameters of the model and the landmark locations of the input image. This finding indicates that one can directly manipulate landmark locations using the gradient of the prediction with respect to the input image.  

Based on this idea, we introduced a fast method of manipulating landmark locations through spatial transformation, which is approximately 200 times faster than the previous geometric attacks, with the success rate of 99.86\%. In addition, we developed a second attack constrained on the semantic structure of the face. The second attack is extremely powerful in generating natural-looking samples that are hard to detect even for the state-of-the-art defense methods.

{\small
\bibliographystyle{ieee}
\bibliography{egbib}
}

\end{document}